%% file: icps2019.tex
\documentclass[conference, a4paper]{IEEEtran}
\IEEEoverridecommandlockouts
\usepackage{mathtools}
\usepackage{amssymb,amsmath}
\usepackage{ifpdf}
\usepackage{url}
\usepackage{graphicx}
\usepackage{caption}
\usepackage{subcaption}
\usepackage{cite}
\usepackage{etoolbox}
\usepackage{multirow}
\usepackage{tabu}
\usepackage{wrapfig}

\usepackage{array}
\usepackage{hyperref}
\newcolumntype{L}[1]{>{\raggedright\let\newline\\\arraybackslash\hspace{0pt}}m{#1}}
\newcolumntype{C}[1]{>{\centering\let\newline\\\arraybackslash\hspace{0pt}}m{#1}}
\newcolumntype{R}[1]{>{\raggedleft\let\newline\\\arraybackslash\hspace{0pt}}m{#1}}

\begin{document}
\nocite{*}
\title{Rare event failure test case generation in Learning-Enabled-Controllers
\thanks{}
}

\author{\IEEEauthorblockN{\textit{Harsh Vardhan,Janos Sztipanovits}}
\IEEEauthorblockA{\textit{{Vanderbilt University}
}
}}
\makeatletter
\patchcmd{\@maketitle}
  {\addvspace{0.5\baselineskip}\egroup}
  {\addvspace{-1\baselineskip}\egroup}
  {}
  {}
\makeatother
\maketitle

\input{0.abstract.tex}
\input{1.Introduction.tex}

\input{3.Approach.tex}
\input{4.Experiments.tex}

\input{7.LearningAndExperimentDetails}
\input{2.RelatedWorks.tex}

\input{6.ConclusionFutureWork.tex}

\bibliographystyle{IEEEtran}
\bibliography{references}

\end{document}

%% file: 0.abstract.tex
\subsection*{\textbf{\textit{ABSTRACT}}}
\label{sec:abstract}
\textbf{
Machine learning models have prevalent applications in many real-world problems, which increases the importance of correctness in the behaviour of these trained models. Finding a good test case that can reveal the potential failure in these trained systems can help to retrain these models to increase the correctness. For a well-trained model, the occurrence of a failure is rare. Consequently, searching these rare scenarios by evaluating each sample in input search space or randomized search would be costly and sometimes intractable due to large search space, limited computational resources, and available time. 
In this paper, we tried to address this challenge of finding these failure scenarios faster than traditional randomized search.  The central idea of our approach is to separate the input data space in \textit{region of high failure probability} and \textit{region of low/minimal failure probability} based on the observation made by training data, data drawn from real-world statistics, and knowledge from a domain expert. Using these information, we can design a generative model from which we can generate scenarios that have a high likelihood to reveal the potential failure.  We evaluated this approach on two different experimental scenarios and able to speed up the discovery of such failures a thousand-fold faster than the traditional randomized search. Source code of experimentation can be found at \href{https://github.com/vardhah/final_codes/tree/master/rare_event_failure}{here.}
}
\subsection*{Keywords}
\textbf{\textit{ Machine learning, failure test scenarios, reinforcement learning, generative model, rare event failure.}}

%% file: 1.Introduction.tex
\section{Introduction}
\setlength{\parskip}{0pt}
\setlength{\parsep}{0pt}
\setlength{\headsep}{0pt}
\setlength{\topskip}{0pt}
\setlength{\topmargin}{0pt}
\setlength{\topsep}{0pt}
\setlength{\partopsep}{0pt}
\label{sec:introduction}

Testing of machine learning models is an emerging and challenging research topic. There are various aspects against which a machine learning model should be tested, like correctness, robustness, security, interpretability etc. In this paper, the scope of testing is in context of \textit{correctness}. Empirical correctness of a machine learning model is statistical quantification of producing desired outcome during prediction.  The goal of testing is to evaluate correctness of a trained model and find a \textit{good test scenario}. We define \textit{good test scenario} as test case that can expose the potential fault in the model behaviour and provide reasoning about correctness.
General software testing paradigm can not be deployed for machine learning models as these are more statistically oriented data-driven programmable entity, where the logical decision boundary is outlined via training process. The behaviour evolves during training process and the result of training cannot be outlined prior to empirical testing, which is contradictory with traditional software testing paradigm, where desired behaviour is fixed first and then underlying behaviour of a software is designed. This behaviour is static in nature and do not dynamically alter with the amount of information in data. Contrary to traditional software testing, test case generation in machine learning is based on observed data. 

Most machine learning models operates in high dimensional space, where sufficient testing by sampling the entire data space is not possible due to enormous number of required samples.  With limited computational resources or available time, the alternative approach is generate input test case by vanilla monte carlo sampling up a maximum up to training iteration or size of training data for detection of failure. Such testing may miss failures entirely, leading to the deployment of unsafe trained model. \textit{Rare event failure} is those failure that have very low probability to occur, and the model performs its desirable behavior most of the time. Finding these scenarios may help to retrain these models and increase the correctness of the model. A systematic and extensively automatable way to search a good test scenario may increase the effectiveness and efficiency of the test and thus to reduce the cost of testing as well as increase correctness in system.

Some early work for searching a good test case in autonomous system uses a fitness function, that assign numerical qualitative value to a test case\cite{wegener2004evaluation}. By defining a fitness function, the search problem can be converted into an optimisation problem, which is mathematically and computationally tractable. The fitness function framework is extended for generating test case in context of machine learning model. 
\cite{hauer2019fitness} addressed the methodological challenge in creating suitable fitness functions by formulating a fitness functions template for testing automated and autonomous driving systems.  By attaching fitness function to different test scenario, recent works uses  either clustering algorithms or evolutionary algorithm to find a good test case for both single and multi-objective fitness function\cite{abdessalem2018testing}\cite{ben2016testing}. 
Another approach to generate a failure test scenario is attempted through adversarial attack\cite{zhang2018deeproad}\cite{zhou2020deepbillboard}. Adversarial input test cases are perturbed version of original inputs so, they may or may not belongs to training data distribution. Although these inputs exposes robustness or security flaws in the trained model, but it is not justified to demand ML model to work on distribution that is not seen during training. 

The goal of this work is expeditiously finding failures scenario in input training data distribution. 
There are only few works in machine learning testing in context of in-distribution rare event failure search. \cite{uesato2018rigorous} implemented a guided search approach to find a rare event failure scenario by learning from training data and focus on region of input space which have high probability to fail based on failures occurred during training. Data from earlier failures was used as a driving signal for segregation of input search space. Using these earlier failures, a failure predictor(called it AVF) was trained, which  produce the likelihood of failure on the given input sample. 
We conducted an independent evaluation of this guided search process in different experimental setting and found that the predictor performance has high variance, which depends on the training process. This underlying reason is  result of non-monotonic improvement in policies during training. As these policies are used for data generation by adding stochasticity or noise to the action for exploration, the convergence to a good policy is arbitrary and earlier failure scenario becomes ineffective to train a good failure predictor. Attempt was also made to control the step-size for
policy updates for making learning smooth, but it had small effect on performance of predictor and fining a good step size for monotonic improvement in policy is not a trivial problem. 

To make search of this rare event process more robust, we extended this approach, by replacing as well as augmenting AVF with a generative model.
The motivation behind using a generative model was: 
\begin{enumerate}
    \item For finding a failure, AVF guided search rely only on training data and training process, and there is no  way to  influence  this  guided  search  by  including  data  derived  by  hand using expert  knowledge,  data from real world collected statistics etc. By replacing AVF with a generative model, that can be trained on same training data as AVF but can also include data from other different sources.
    \item Generative models can be trained in online fashion, where current probability distribution become prior for next phase of training and then this prior can be conditioned over new data (derived from any source). Accordingly, test case generation process adapts with new discoveries and able to find rare event scenarios faster. 
    \item One we have a generative scheme, We can sample form this learned probability distribution function to generates new unseen scenarios which may cause failure and reveal potential faulty system operation.
\end{enumerate}
The generative model guided search approach can be used stand-alone or in augmentation of AVF guided search. In augmentation of AVF guided search, generative model outperforms and dynamically adapt and learn to find these failure scenario faster.  We empirically evaluated all three search methods (AVF guided,Generative model guided and combination of both) by training a car braking system in a simulated environment. Using both kind of guided search, we find failures much faster than traditional Vanilla Monte Carlo approach. 

%% file: 3.Approach.tex
\section{Problem Formulation and Approach}
\setlength{\parskip}{0pt}
\setlength{\parsep}{0pt}
\setlength{\headsep}{0pt}
\setlength{\topskip}{0pt}
\setlength{\topmargin}{0pt}
\setlength{\topsep}{0pt}
\setlength{\partopsep}{0pt}

\setlength{\belowcaptionskip}{-10pt}
\captionsetup{justification=raggedright,singlelinecheck=false}

\label{sec:approach}

For designing a machine learning based controller in reinforcement learning setting, training process involves running simulation experiments with aim to learn a policy for the desired behaviour by continuous interaction with environment. These trained models are generally called an \textit{agent}. 
Before formulating the problem, let's introduce some notations which we will use in rest of the paper.  
The state information $X$ represent agent's state in the environment, which agent perceive from its environment. $Z$ represent the environment's state (which may be environment's private representation), which is sampled from some unknown distribution $\mathbb{P}_Z$. 
In our experimental setting, experimenter neither observe nor control $Z$. 
$X$ is drawn from some distribution $\mathbb{P}_X$, which do not need to be known. 
Only requirement we impose on $X$ is, experimenter should be able to sample it quickly with approximately no cost in comparison to running the simulation.
Our interest is agent’s performance on the environment distribution ($\mathbb{P}_Z$) over initial conditions $X \sim \mathbb{P}_X$.
Once training is done, we assess the trained agent by rolling out experiments with the trained agent given an initial condition $X \sim \mathbb{P}_X$.
The outcome of experiment can be failure or success, which is a random variable called \textit{failure indicator} $C$. Here, $C$ has binary value and $C = 1$ indicates a “failure”. 
We are interested here to find a good test case, which can reveal the potential fault in trained controller. The problem can be formulated as finding those $X \sim \mathbb{P}_X$ on which the trained controller fails.

\begin{figure}[h!]
        \centering
        \captionsetup{justification=centering}
        \includegraphics[width=0.45\textwidth]{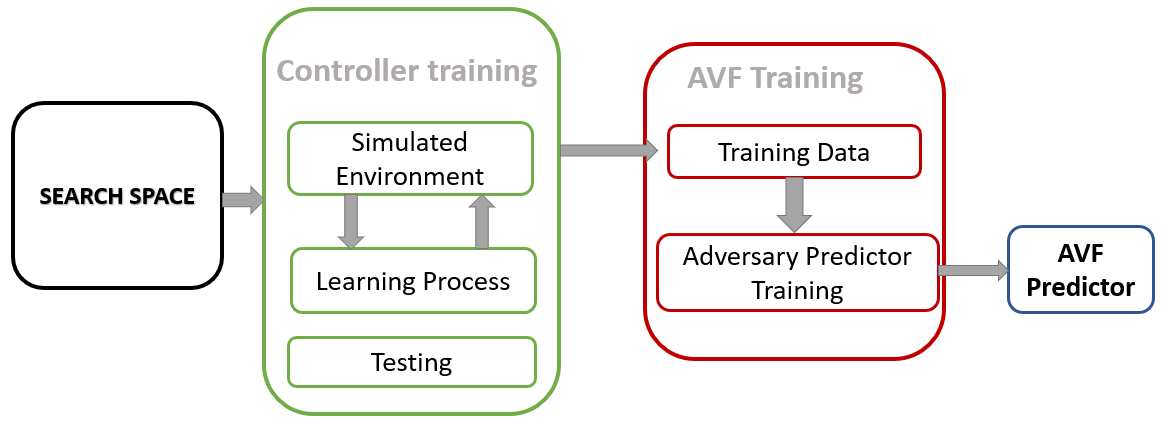}
        \caption{Phases of Training}
        \label{fig:phases}
    \end{figure}

The naive approach to find a failure ,vanilla monte carlo may not find a failure even after evaluating for same number of episode as training episodes. It may give a sense of safe controller, which may or may not true. 
To make evaluation more rigorous, one approach is by using a priority replay (PR) adversary search, which test the trained model on all earlier failures that happened during training. Generally a well trained agent do not fail during priority replay search as it has learnt from all failure occurred during training. 
After failing to find failure during PR adversary search, it is obvious to switch to randomized search (monte carlo sampling) from input search space and run simulation to observe the results. System with large input search space, especially in case of all CPS which works in continuous domain,finding a failure using monte carlo is really expensive in terms of number of simulation episodes required to find a failure case. 
To speed up the search of failures in learned agent, \cite{uesato2018rigorous} proposed AVF (failure probability predictor) guided search. The motivation behind AVF guided search is to screen out situations that are unlikely to be problematic, and focus evaluation on the most difficult situations (corner cases) in probabilistic rigorous framework. 
AVF learns the failure pattern from earlier agent’s failure. During testing of agent, it assign probability of failure for the each input search space. It is advantageous as if sampling from search space and its prediction using AVF is computationally cheaper, we can narrow our search of failure scenarios to regions, which has high probability of failure, and save on simulation's computation. Given an input data, AVF predicts probability of failure conditioned on observed training data.
Finding failure using AVF involves two phase of training \textit{controller training} and \textit{AVF training} (refer fig \ref{fig:phases}). In controller training phase, we train a model using some learning method to achieve a learning goal. As training progress, the learned controller becoming more robust and will have less and less failure. At each episode during training, we specify initial conditions $X \sim \mathbb{P}_X$ and observe the outcome $C\,\to\,\{0,1\}$. For any episodic simulation, it is guaranteed to get a return value of $C$. During training, initial condition($X$), simulation episodic outcome($C$) and information about agent($\theta$) is collected. Here, $\theta$ encodes all information that can tell about agent state (like stochastic noise added during each episode or episode number). So, at each episode of training $t$, we start simulation with $X_t$, observe $C_t$ and $\theta_t$. The training data in form of \{$(X_1; \theta_1; C_1),(X_2; \theta_2; C_2), \cdots , (X_n ; \theta_n; C_n )$\} will be used to train AVF in AVF training phase. 
Once training is finished, controller is tested up to maximum of training episode. If testing do not give us any failure, we assume our trained model is functionally robust, which may or may not be true.
\\
During AVF training phase, we used training data to train a neural network in supervised setting. In initial phase of training, agent is prone to fail more often so,these failure data from initial phase of training is ignored and only training data from last phase of training is used for training AVF. $X$ and $\theta$ form the input feature space and $C$ would be the output of neural network. Neural network is trained to returns the probability of failure given a initial condition and information about agent. $\mathcal{NN}:(X,\theta)\,\to\,[0,1]$.

If $f_*$ is the real failure probability function given initial state $X$ and $\theta$, the goal of training AVF ($\mathcal{NN}$) is to approximate $f_*$ up to normalisation constant.  
 \begin{gather*}
  f_*(x)= \mathbb{P}(C=1|{X,\theta}) ;  \quad  X \sim P_X \\
  \mathcal{NN} \approx f_* 
 \end{gather*}

\begin{figure}[h!]
        \centering
        \captionsetup{justification=centering}
        \includegraphics[width=0.45\textwidth]{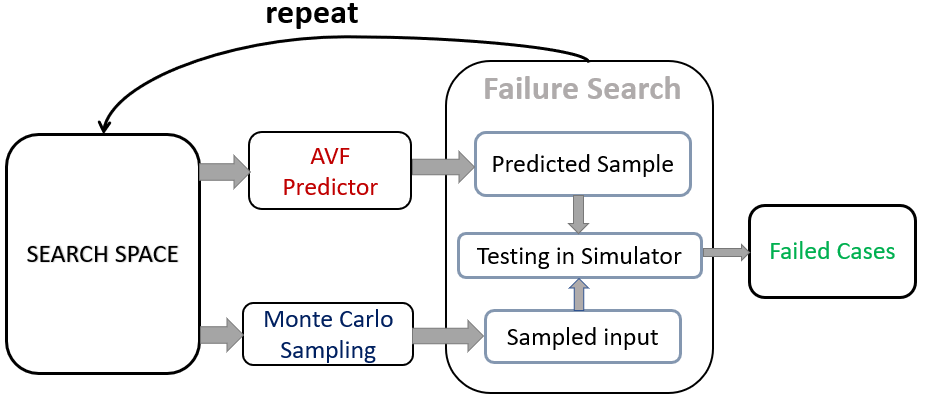}
        \caption{Failure search - AVF and Monte Carlo}
        \label{fig:VMC_AVF}
    \end{figure} 
In search of a failure condition, the traditional \textit{Vanilla Monte Carlo(VMC)} sampling samples $X$ from $\mathbb{P}_X$ and run the experiment. If experiment is episodic , then result of simulation will be success($C=0$) or failure($C=1$). For a well trained agent, finding a failure case may take way more number of episodes than training episodes. 
In contrast, when using AVF guided search for finding a failure, obvious approach is to evaluate the agent on
${argmax}\mathcal{NN}$(x). 
However, maximization using neural network is not feasible and also we only have knowledge of approximate probability, so a feasible approach is sample $n$ initial conditions from $\mathbb{P}_X$, pick the initial condition from this set where $\mathcal{NN}$ is the largest, and run the experiment from the found initial condition. We repeat this process until we find a failure. The reasoning behind predicting first through $\mathcal{NN}$ is to discard all the search space samples which have very low probability of failure and select the one who has greater chance of failure. One assumption is implicit that cost of evaluating a sample on $\mathcal{NN}$ is negligible in comparison to running an experiment, then only AVF guided search is advisable in comparison to VMC. The idea is to use NN to save on the cost of experimentation.

This guided search method has some limitations that we observed during empirical evaluation and also have scope of improvement. The first limitation of AVF guided search is to rely completely on training data and training process for searching for a failure. The training process in reinforcement learning is not guaranteed to be monotonic. Research efforts like constraining gradient updates of policy parameters(TRPO)\cite{schulman2015trust}, stochastic weight averaging (SWA)\cite{nikishin2018improving} etc attempts to learn a monotonically increasing control policy, but there is no guarantee for such convergence. Second limitation of AVF guided search is it does not adapt with discoveries of new failed cases, in such case we donot improve this search process. In large state space, the adaptive search process is beneficial by saving cost and time by adapting with new discoveries. Third limitation is, there is no way to influence this guided search by including data derived using expert knowledge, statistics collected in real world or other simulators and similar experiments, etc that may help to find more realistic and speedier discovery of test scenarios. 

To tackle all above mentioned issues, in this paper attempt is made to learn a  probability distribution of the observed failures. If $X_{f1},X_{f2}, \cdots , X_{fn}$ are collection of failure instances, the goal is to model $\mathbb{P}(X_f)$. This trained probability distribution tries to imitate and approximate the real failure probability distribution. We can also sample from this distribution to generate a test scenario, which has high probability to the failure. This generative model may work stand alone or in augmentation of AVF. Empirically it is observed that in augmentation with AVF, it can predict good test scenario faster than AVF.
It can be trained online i.e. once we obtain some new failure scenario data, the old posterior becomes the new prior and then we condition our prior using Bayes rule to get new posterior (i.e. new probability distribution). We used mixture model-based clustering algorithm to model the $\mathbb{P}(X_f)$. We choose a parametric probability distribution model $\mathbb{P}(X_f | w)$ to fit our data distribution by finding the best parameter(w) which represent $\mathbb{P}(X_f)$, where $w$ is the parameter of model. 
\begin{figure}[h!]
        \centering
        \captionsetup{justification=centering}
        \includegraphics[width=0.49\textwidth]{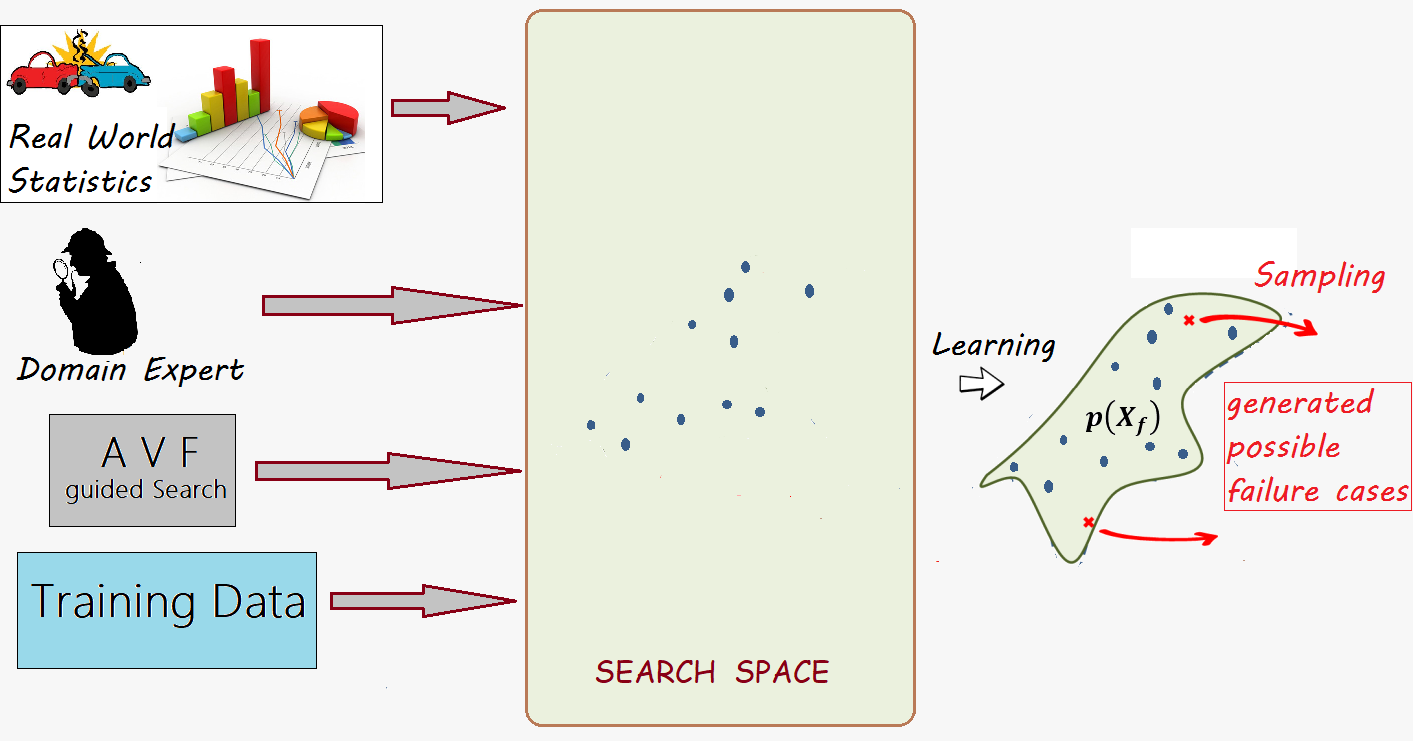}
        \caption{Motivation behind generative model guided search}
        \label{fig:Gen_AVF}
    \end{figure} 
Gaussian Mixture Model\cite{GMM}, which is a popular mixture model is reasonable model to work with. So, the problem of modeling probability distribution of data is now training a GMM which can be written as: 
\begin{gather*}
    \mathbb{P}(X_f |w)= \pi_1*\mathcal{N}(X_f|\mu_1,\Sigma_1)+ \cdots + \pi_n*\mathcal{N}(X_f|\mu_n,\Sigma_n) \\
    w=\{\mu_1,\Sigma_1,\pi_1,\cdots, \mu_n,\Sigma_n\}
\end{gather*}
Here $n$ is a hyper-parameter, which depends on distribution structure of data and can be tuned using Bayesian Information criterion and cross-validation.

\begin{figure}[h!]
        \centering
        \captionsetup{justification=centering}
        \includegraphics[width=0.45\textwidth]{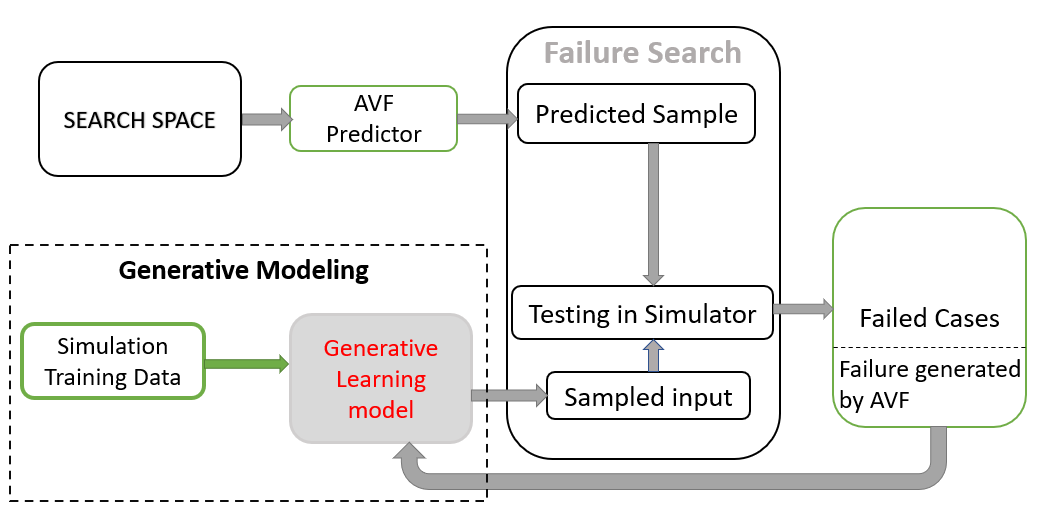}
        \caption{Failure search - AVF in augmentation with Generative model}
        \label{fig:Gen_AVF}
    \end{figure} 
Training data that was used for AVF training can be reasonable starting point for modeling this distribution, once we have more and more data from various sources, we can retrain the model to make it better and better. 
After training, we can sample from this family of multivariate normal distribution for a probable test candidate. We run the simulation experiment with this sampled initial condition($X$) and observe the failure/success($C$). Trained using earlier failure cases and other data sources, there is high probability to generate samples which can fail and very low probability of generating data in the region of search space which are less likely to fail. 
This Generative model can also work in augmentation of AVF guided search, where failure cases generated by AVF guided search also contribute to dynamically modify the probability distribution model (refer fig \ref{fig:Gen_AVF}).

%% file: 4.Experiments.tex
\section{Experiments}
\setlength{\parskip}{0pt}
\setlength{\parsep}{0pt}
\setlength{\headsep}{0pt}
\setlength{\topskip}{0pt}
\setlength{\topmargin}{0pt}
\setlength{\topsep}{0pt}
\setlength{\partopsep}{0pt}
\captionsetup{justification=raggedright,singlelinecheck=false}

\label{sec:experimentalResults}
To empirically evaluate above mentioned approaches, We set up two experiments. In both cases,we used reinforcement learning approach called DDPG(Deep Deterministic Policy Gradient)\cite{lillicrap2015continuous} based learning model to design the braking system of a car. In \textit{Scenario 1}, an approaching car detects an obstacle at distance of 100 meters and the learning goal is stop without crashing. The static and kinetic friction coefficient of road is constant throughout the experiment. The random variable in this scenario is speed of the car when it detects the obstacle, which is drawn from a normal distribution with mean velocity 38 miles/hour and standard deviation 11 miles/hour. $Initial\_Speed(v) \sim \mathcal{N}(38,11)$ 
    \begin{figure}[h!]
        \centering
        \captionsetup{justification=centering}
        \includegraphics[width=0.45\textwidth]{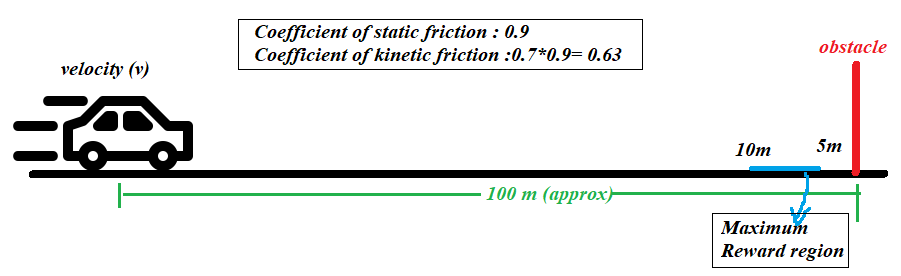}
        \caption{Scenario 1}
        \label{fig:exp1}
    \end{figure}
    
In \textit{Scenario 2}, to make environment more realistic and complex, we introduce a patch on the road having varying friction coefficients, size and location. This increases complexity of learning problem and randomness in agent perceived state to higher dimension ($4$ dimensions). Speed of the car when it detects the obstacle in this scenario is drawn from a normal distribution with mean velocity of 35 miles/hour and standard deviation of 9 miles/hour. $Initial\_Speed \sim \mathcal{N}(35,9)$ 
 
%
       \begin{figure}[h!]
        \centering
        \captionsetup{justification=centering}
        \includegraphics[width=0.45\textwidth]{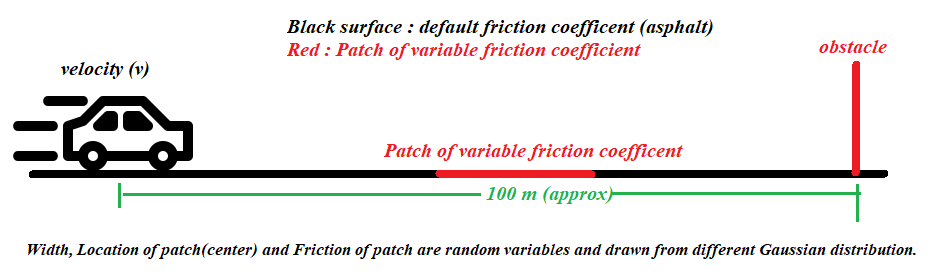}
        \caption{Scenario 2}
        \label{fig:exp2}
    \end{figure} 
Learning goal in both cases is to learn a controller for braking. The reward setting is such that the vehicle when brakes around region of 5-10 meters gets the maximum rewards.

We will compare Vanilla Monte Carlo based adversaries search with AVF guided search and GMM guided search by the number of episodes required to find an initial condition that results in a trajectory that ends with a failure. Our purpose is to illustrate three key points.
First,VMC guided search, even when using the same number of episodes as training the agent, can lead to a false sense of safety by failing to detect any catastrophic failures. Second, AVF guided search
addresses this issue and reducing the cost of finding failures to large extent. Third, generative model guided search also does better than VMC guided search and in augmentation with AVF it is the most flexible and dynamic approach to find a failure.

\begin{table}
\centering
\normalsize
\begin{tabular}{  L{1.5 cm}  L{1.8cm} L{1.8 cm} L{1.8 cm} } 
\hline
\textbf{Failure number} & \textbf{VMC} & \textbf{ AVF } & \textbf{ GMM } \\
\hline
1 & 16349 & 13  & 1 \\
\hline
2 & 518 & 1 & 1  \\
\hline
3 & 131188 & 66 & 8  \\
\hline
4 & 8217 & 2 & 13 \\
\hline
5 & 28033 & 20 & 7  \\
\hline
6 & 25481 & 11 & 1  \\
\hline
7 & 21184 & 24 & 5   \\
\hline
8 & 26979 & 41  & 3 \\
\hline
9 & 8566 & 49 & 15 \\
\hline
10 & 36466 & 16 & 2\\
\hline
\textbf{Average} & \textbf{30298} & \textbf{24} & \textbf{6}\\
\hline
\end{tabular}
\captionsetup{justification=centering,labelsep=period}
\caption{Episodes of simulation for 10 consecutive failure search in \textit{Scenario2}. Agent is trained for 15000 episodes. Average number of episodes required in case of VMC is almost double to training episodes and way higher than AVF and GMM guided search }
\label{tab:e10}
\end{table}

Table \ref{tab:e10} shows the number of simulation episodes rolled out in search of ten different failure cases in \textit{Scenario 2}. In average case, the VMC takes 30298 episodes, which is almost double to number of training episodes. In such case, if agent is tested  only up to maximum of number of training episodes, it has high probability to give a false sense about robustness of this trained system. However, AVF took only 6 episodes for finding a failure. So, VMC based adversary search is not only inefficient , it may also give lead to deploy an agent which may result in catastrophic failure.

Table \ref{tab:failures} shows number of episodes rolled out required in minimum, maximum and average case in search of 100 failures for both scenario 1 and scenario 2. We also test GMM guided search and GMM in augmentation with AVF guided search. For GMM and GMM+AVF, we didnot used data from domain expert knowledge and real world statistics. However, including these data will surely increase the probability of getting better failure cases. For GMM , we used the same data that we used for AVF training. However while using GMM+AVF, we first trained GMM and during failure search process, we sample one test case from both GMM and AVF alternatively. If we get any failure from sample generated by AVF we modify our probability distribution with this new observed data. For \textit{scenario1}, in average case, it took 10738 episodes which is almost double of number of training episodes. AVF guided search took only 6 episode to find a failure in average case. For \textit{scenario2}, in average case, agent took 22252 episodes which is almost 1.5 times to number of training episodes. AVF guided search took only 36 episode to find a failure in average case. AVF guided search does way better than Vanilla Monte Carlo search.

In average case, GMM guided search does better than VMC in both scenarios.  However in \textit{Scenario2}, GMM search is faster than AVF guided search. This is because of kind of training data generated during training. If training data used of AVF/GMM training is far from real failure data in search space then GMM perform poorer. However it will still does better than VMC. Our motivation for using GMM model is to reduce dependency only on training data for generating a failure and incorporate data from other sources like domain expert knowledge , real world collected statistics which is not possible in case of AVF guided search. 

GMM in augmentation of AVF perform twice faster than AVF only guided search in \textit{scenario1} and five times faster than AVF only in \textit{scenario2}. AVF with GMM is the most flexible and dynamic approach to find faioure in learning enable system.

\begin{table*}
\centering
\normalsize
\begin{tabular*}{\textwidth}{c @{\extracolsep{\fill}} ccccc}
\hline
\textbf{Scenario} & \textbf{VMC} & \textbf{ AVF } & \textbf{ GMM } & \textbf{ GMM +AVF}\\
\hline
\textit{Scenario1} & 82/10738/47323 & 1/6/34  & 4/3991/21393 & 1/3/26 \\
\hline
\textit{Scenario2} & 3/22252/154374 & 1/36/102 & 1/4/19 & 1/5/27\\
\hline

\end{tabular*}
\captionsetup{justification=centering,labelsep=period}
\caption{ Cost for different search approaches to find a failure case, measured by the number of simulation episodes (for 100 failures). Each column reports the Min/Average/Max number of simulation episodes for 100 failures . In \textit{Scenario1}, AVF+GMM guided search is two times faster than AVF guided search in average case. However GMM guided search, does worse in scenario 1 than in scenario 2, but it is still faster than VMC. \textbf{Training Episodes}: \textit{Scenario1}(5000) ; \textit{Scenario2}(15000)}
\label{tab:failures}
\end{table*}

%% file: 7.LearningAndExperimentDetails.tex
\section{Training and Experiment Details}
\setlength{\parskip}{0pt}
\setlength{\parsep}{0pt}
\setlength{\headsep}{0pt}
\setlength{\topskip}{0pt}
\setlength{\topmargin}{0pt}
\setlength{\topsep}{0pt}
\setlength{\partopsep}{0pt}
\captionsetup{justification=raggedright,singlelinecheck=false}

\label{sec:ledetails}

\subsection{Agent Training Details :}
In \textit{scenario 1}, at each step agent receives a 3-dimensional observation vector summarizing its position, friction coefficient of road and velocity of car. 
We set up two type of rewards, \textit{immediate reward}, which agent receives after each action it takes and \textit{terminal reward}, which agent receive at the end of episode. Immediate reward is small in comparison to terminal reward. In theory, the learning goal can be achieved by only using terminal rewards, but empirically observation suggests sparse reward( by just using terminal rewards) make learning difficult. 
Immediate reward that agent receives is reward proportional to its change in velocity. Terminal reward penalises/rewards the agent based on where it stops. We want vehicle to stop to the maximum reward region(5-10m from obstacle). Episode is terminated if vehicle is either stopped or collided with obstacle. Collision with obstacle provides a large negative reward. 
 In scenario 2, at each step agent receives a 4-dimensional observation vector summarizing its position, velocity, friction coefficient of road and the position of  at the vehicle's position. 
Problem instances are sampled by randomly sampling an initial velocity from a fixed distribution.

In both scenarios, we used an batched implementation of Actor-Critic, using a
Deep Deterministic Policy Gradient (DDPG) algorithm \cite{lillicrap2015continuous}. For scenario 1, we trained agent for 5000 episodes and for scenario2, we trained agent for $15K$ episodes. Output layer of the actor is a single neuron with sigmoid activation function which gives brake output in range $[0,1]$.  For exploration purpose, We used setting used in original DDPG paper\cite{lillicrap2015continuous} by adding stochastic noise in the output drawn from a  Ornstein–Uhlenbeck(OU) process with $\mu$= 0.2, $\theta$= 1, $\sigma$= 0.1.

\subsection{Experimental Details :}
\subsubsection{AVF}
For training of AVF we ignored failure data generated by agents during early phase of training as agent fails more frequently . For \textit{Scenario 1}, we used last 300 episodes of training data and for \textit{Scenario  2}, we used last 5000 episodes of training data. 
For including information about agent , we included training episode. These data is concatenated and normalised before using for training of AVF. The architecture used for AVF training is a three layer neural network with 64 (Layer1,  relu), 32 (Layer2, relu) and 1 (Output layer, Sigmoid) neurons. The learning goal to minimize the binary cross-entropy loss, with the Adam optimizer \cite{kingma2014adam}. Training is done 
for 500 epoch in both scenario, which took 4-5
minutes on a single GPU. 

\subsubsection{Generative model} We used same training data that is used for AVF to model our data distribution. We used multivariate Gaussian mixture model(GMM) to model our data distribution. 
\begin{gather*}
    \mathbb{P}(X_f |w)= \pi_1*\mathcal{N}(X_f|\mu_1,\Sigma_1)+ \cdots + \pi_n*\mathcal{N}(X_f|\mu_n,\Sigma_n) \\
    w=\{\mu_1,\Sigma_1,\pi_1,\cdots, \mu_n,\Sigma_n\}
\end{gather*}
The training problem in this case would be 
\begin{gather*}
    \underset{w}{max} \; \prod_{i=1}^{N} P(X_f |w) =
    \prod_{i=1}^{N}(\pi_1*\mathcal{N}(X_f|\mu_1,\Sigma_1)+ \cdots ) \\
    subjected\; to\; \pi_1+\cdots+\pi_n =1 ; \pi_k \geqslant0;\; k=1,\cdots n\\
    \Sigma_k >0 \; k=1,\cdots n
\end{gather*}
The first hyper parameter to tune is the number of mixture model($n$). We did hyper-parameter tuning and found n=2 for \textit{Scenario1} and n=3 for \textit{Scenario2} based on Bayesian Information Criterion and cross-validation. We used Expectation-Maximisation(EM) algorithm\cite{bishop2006pattern} with full co-variance matrix to train the GMM. EM is iterative process to find maximum a posteriori (MAP) estimates of parameter($w$) , where the data is modelled on unobserved latent variables. EM also depends on the initialisation value of $w$, we made 100 different initialisation and selected one which represent data the most. 

%% file: 2.RelatedWorks.tex
\section{Related Work}
\setlength{\parskip}{0pt}
\setlength{\parsep}{0pt}
\setlength{\headsep}{0pt}
\setlength{\topskip}{0pt}
\setlength{\topmargin}{0pt}
\setlength{\topsep}{0pt}
\setlength{\partopsep}{0pt}

\label{sec:relatedWorks}
 \cite{szegedy2013intriguing} proposed several adversarial example generation methods for attacking Deep Neural Networks. These methods majorly generated an adversarial example via adding calibrated perturbation to an image data that can confuse
the DNN.\cite{goodfellow2014generative} \cite{moosavi2016deepfool} have shown various techniques to generate adversarial examples in Deep Neural network(DNN). All of these work are done for image data. We use the the similar DNN to learn a policy in reinforcement learning setting and consequently they fail in expected way. 

 Some early work in finding failure in RL agent is done by \cite{huang2017adversarial} \cite{lin2017tactics}. \cite{huang2017adversarial} proposes an
adversarial attack tactic where the adversary attacks a deepRL agent at every time step in an episode. To create failure, it considers adversaries have capability to  introducing small perturbations to the raw input of the DNN. They added different perturbation based on whether the adversary has access to the Deep policy network or not.
\cite{lin2017tactics} modified this approach on attempting to find the failure of RL agent. For that they designed two different attacks on the trained agent. In one case, they  aimed to minimizing the agent’s reward by attacking the agent in some intermittent simulation time steps in an episode. In another form of attack they lured the agent to some designated state rather than reaching to target state. In this kind of attack they predicted the future state by using generative model and generated a sequence of actions which divert agent from its target position. 

Both \cite{lin2017tactics} \cite{huang2017adversarial}  tried to find failure in a trained agent but in totally different context. In their context they generated adversarial situation completely out of training distribution and it is more likely that in such circumstances these DNN can be failed easily. Our work can be complementary to it as it try to find failure scenario within the training distribution.  While in our case we

%% file: 6.ConclusionFutureWork.tex
\section{Conclusion}
\setlength{\parskip}{0pt}
\setlength{\parsep}{0pt}
\setlength{\headsep}{0pt}
\setlength{\topskip}{0pt}
\setlength{\topmargin}{0pt}
\setlength{\topsep}{0pt}
\setlength{\partopsep}{0pt}
\captionsetup{justification=raggedright,singlelinecheck=false}
\label{sec:conclusionFutureWork}
In this work, we showed that standard approaches to evaluating failure in machine learning models are highly inefficient in
detecting rare failure cases. In such case, there is possibility to deploy a trained model which will give an illusion of completeness and safety. To increase the robustness, we can draw knowledge from our training experience and other sources to find the scenarios that may result into failure much faster than using traditional methods. In future work we will be assessing this method on other machine learning approaches like supervised learning as well as real world scenarios and data.